\documentclass[11pt,a4paper]{article}
\usepackage[hyperref]{naaclhlt2018}
\usepackage{times}
\usepackage{latexsym}
\usepackage{booktabs}
\usepackage{graphicx}
\usepackage{siunitx}
\usepackage{graphics}
\usepackage{amsmath,amssymb}
\usepackage{multirow}
\usepackage{comment}

\usepackage{caption}
\usepackage{url}
\usepackage{makecell}
\usepackage{array}

\aclfinalcopy 
\def\aclpaperid{1731} 

\title{Can Neural Machine Translation be Improved with User Feedback?}

\author{Julia Kreutzer$^{1}$\thanks{\;The work for this paper was done while the first author was an intern at eBay.} \and Shahram Khadivi$^{3}$ \and Evgeny Matusov$^{3}$ \and Stefan Riezler$^{1,2}$ \\
$^1$Computational Linguistics \& $^2$IWR, Heidelberg University, Germany \\
{\tt \small \{kreutzer,riezler\}@cl.uni-heidelberg.de} \\
$^3$eBay Inc., Aachen, Germany \\
{\tt \small \{skhadivi,ematusov\}@ebay.com}
}

\date{}

\begin{document}
\maketitle
\begin{abstract}
We present the first real-world application of methods for improving neural machine translation (NMT) with human reinforcement, based on explicit and implicit user feedback collected on the eBay e-commerce platform. Previous work has been confined to simulation experiments, whereas in this paper we work with real logged feedback for offline bandit learning of NMT parameters. We conduct a thorough analysis of the available explicit user judgments---five-star ratings of translation quality---and show that they are not reliable enough to yield significant improvements in bandit learning. In contrast, we successfully utilize implicit task-based feedback collected in a cross-lingual search task to improve task-specific and machine translation quality metrics.
\end{abstract}

\section{Introduction}

In commercial scenarios of neural machine translation (NMT), the one-best translation of a text is shown to multiple users who can reinforce high-quality (or penalize low-quality) translations by explicit feedback (e.g., on a Likert scale) or implicit feedback (by clicking on a translated page). In such settings this type of feedback can be easily collected in large amounts. While bandit feedback\footnote{The fact that only feedback for a single translation is collected constitutes the ``bandit feedback'' scenario where the name is inspired by ``one-armed bandit'' slot machines.} in form of user clicks on displayed ads is the standard learning signal for response prediction in online advertising \cite{BottouETAL:13}, bandit learning for machine translation has so far been restricted to simulation experiments \cite{sokolovNIPS2016,lawrence-sokolov-riezler:2017:EMNLP2017,nguyen-daumeiii-boydgraber:2017:EMNLP2017,kreutzer-sokolov-riezler:2017:Long,bahdanau2017actor}.

The goal of our work is to show that the gold mine of cheap and abundant real-world human bandit feedback can be exploited successfully for machine learning in NMT. We analyze and utilize human reinforcements that have been collected from users of the eBay e-commerce platform. We show that explicit user judgments in form of five-star ratings are not reliable and do not lead to downstream BLEU improvements in bandit learning. In contrast, we find that implicit task-based feedback that has been gathered in a cross-lingual search task can be used successfully to improve task-specific metrics and BLEU. 

Another crucial difference of our work to previous research is the fact that we assume a counterfactual learning scenario where human feedback has been given to a historic system different from the target system. Learning is done offline from logged data, which is desirable in commercial settings where system updates need to be tested before deployment and the risk of showing inferior translations to users needs to be avoided. Our offline learning algorithms range from a simple bandit-to-supervised conversion (i.e., using translations with good feedback for supervised tuning) to transferring the counterfactual learning techniques presented by \citet{lawrence-sokolov-riezler:2017:EMNLP2017} from statistical machine translation (SMT) to NMT models. To our surprise, the bandit-to-supervised conversion proved to be very hard to beat, despite theoretical indications of poor generalization for exploration-free learning from logged data \cite{LangfordETAL:08,StrehlETAL:10}. However, we show that we can further improve over this method by computing a task-specific reward scoring function, resulting in significant improvements in both BLEU and in task-specific metrics.

\section{Related Work}\label{sec:related}

\citet{sokolov-EtAl:2016:P16-1, sokolovNIPS2016} introduced learning from bandit feedback for SMT models in an interactive online learning scenario: the MT model receives a source sentence from the user, provides a translation, receives feedback from the user for this translation, and performs a stochastic gradient update proportional to the feedback quality. \citet{kreutzer-sokolov-riezler:2017:Long} showed that the objectives proposed for log-linear models can be transferred to neural sequence learning and found that standard control variate techniques do not only reduce variance but also help to produce best BLEU results. 
\citet{nguyen-daumeiii-boydgraber:2017:EMNLP2017} proposed a very similar approach using a learned word-based critic in an advantage actor-critic reinforcement learning framework. 
A comparison of current approaches was recently performed in a shared task where participants had to build translation models that learn from the interaction with a service that provided e-commerce product descriptions and feedback for submitted translations \cite{sokolov-EtAl:2017:WMT}. 
\citet{lawrence-sokolov-riezler:2017:EMNLP2017,LawrenceETALnips:17} were the first to address the more realistic problem of offline learning from logged bandit feedback, with special attention to the problem of exploration-free deterministic logging as is done in commercial MT systems. They show that variance reduction techniques used in counterfactual bandit learning \cite{dudik2011doubly,BottouETAL:13} and off-policy reinforcement learning \cite{PrecupETAL:00,JiangLi:16} can be used to avoid degenerate behavior of estimators under deterministic logging. 

\section{User Feedback}\label{sec:feedback}

\subsection{Explicit Feedback via Star Ratings} \label{sec:explicit}
\begin{figure}[t]
\includegraphics[width=\columnwidth]{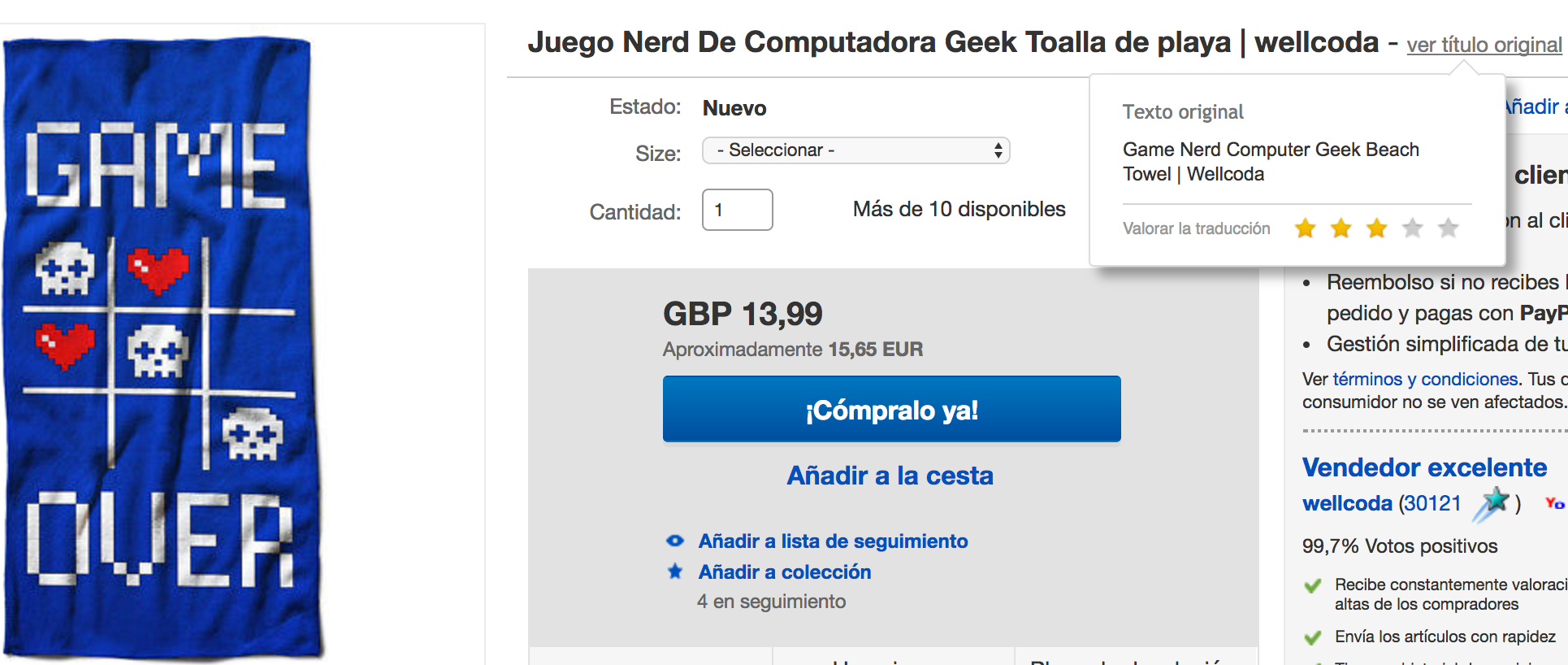}
\caption{Screenshot of the 5-star rating interface for a product on \url{www.ebay.es} translated from English to Spanish.}
\label{fig:interface}
\end{figure}

One way to collect reinforcement signals from human users of the eBay platform is by explicit ratings of product title translations on a five-point Likert scale. More specifically, when users visit product pages with translated titles, they can inspect the source when hovering with the mouse over the title. Then five stars are shown with the instruction to `rate this translation'. A screenshot of an implementation of this rating interface is shown in Figure \ref{fig:interface}.
The original title, the translation and the given star rating are stored.
For the experiments in this paper, we focus on translations from English to Spanish. The user star rating data set contains 69,412 rated product titles with 148k individual ratings. Since 34\% of the titles were rated more than once, the ratings for each title are averaged. We observe a tendency towards high ratings, in fact one half of the titles are rated with five stars (cf. Appendix \ref{sec:rewards}).

To investigate the reliability and validity of these ratings, we employed three bilingual annotators (`experts') to independently re-evaluate and give five-star ratings for a balanced subset of 1,000 product title translations. The annotators were presented the source title and the machine translation, together with instructions on the task provided in Appendix  \ref{sec:instructions}.
The inter-annotator agreement between experts is relatively low with Fleiss' $\kappa=0.12$ \cite{fleiss1971measuring}. Furthermore, there is no correlation of the averaged `expert' ratings and the averaged user star ratings (Spearman's $\rho=-0.05$). 
However, when we ask another three annotators to indicate whether they agree or disagree with a balanced subset of 2,000 user ratings, they agree with 42.3\% of the ratings (by majority voting). In this binary meta-judgment task, the inter-annotator agreement between experts is moderate with $\kappa=0.45$. 
We observe a strong tendency of the expert annotators to agree with high user ratings and to disagree with low user ratings.
Two examples of user ratings, expert ratings and expert judgment are given in Table \ref{tab:userexample}. In the first example, all raters agree that the translation is good, but in the second example, there is a strong disagreement between users and experts. 

\begin{table*}[t]
\resizebox{\textwidth}{!}{%
\begin{tabular}{c|c|ccc}
\toprule
\thead[cc]{\textbf{Source Title}} & \thead[cc]{\textbf{Title Translation}} & \thead{\textbf{User Rating}\\ \textbf{(avg)}} & \thead{\textbf{Expert Rating}\\ \textbf{(avg)}} & \thead{\textbf{Expert Judgment}\\ \textbf{(majority)}} \\
\midrule
 \makecell[tl]{Universal 4in1 Dual USB Car Charger \\Adapter Voltage DC 5V 3.1A Tester For iPhone} & \makecell[tl]{Coche Cargador Adaptador De Voltaje\\ Probador De Corriente Continua 5V 3.1A para iPhone} & 4.5625 & 4.33 & Correct \\
 \midrule
\makecell[tl]{BEAN BUSH THREE COLOURS: YELLOW\\ BERGGOLD, PURPLE  KING AND GREEN TOP CROP} & \makecell[tl]{Bean Bush tres colores: Amarillo Berggold, p\'{u}rpura\\ y verde Top Crop King} & 1.0  & 4.66 & Incorrect \\
\bottomrule
\end{tabular}%
}
\caption{Examples for averaged five-star user ratings, five-star expert ratings and expert judgments on the user ratings.} 
\label{tab:userexample}
\end{table*}

This analysis shows that it is generally not easy for non-professional users of the e-commerce platform, and even for expert annotators, 
to give star ratings of translations in the domain of user-generated product titles with high reliability.  
This problem is related to low validity, i.e.,  we do not know whether the users' response actually expresses translation quality, since we cannot control the influence of other factors on their judgment, e.g., the displayed image (see Figure \ref{fig:interface}), the product itself, or the users' general satisfaction with the e-commerce transaction, nor can we exclude the possibility that the user judgment is given with an adversarial purpose. Furthermore, we do not have control over the quality of sources\footnote{Most titles consist of a sequence of keywords rather than a fluent sentence. See \citet{calixto-EtAl:2017:EACLshort} for a fluency analysis of product titles.}, nor can we discern to which degree a user rating reflects fluency or adequacy of the translation.

\subsection{Task-Based Implicit Feedback} \label{sec:implicit}

Another form of collecting human reinforcement signals via the eBay e-commerce platform is to embed the feedback collection into a cross-lingual information retrieval task. 
The product title translation system is part of the search interaction of a user with the e-commerce platform in the following way: When a user enters a query in Spanish, it is first translated to English (query translation), then a search engine retrieves a list of matching products, and their titles are translated to Spanish and displayed to the user. As soon as the user clicks on one of the translated titles, we store the original query, the translated query, the source product title and its translation. 
From this collection we filter the cases where (a) the original query and the translated query are the same, or (b) more than 90\% of the words from the query translation are not contained in the retrieved source title. In this way, we attempt to reduce the propagation of errors in query translation and search. This leaves us with a dataset of 164,065 tuples of Spanish queries, English product titles and their Spanish translations (15\% of the original collection). Note that this dataset is more than twice the size of the explicit feedback dataset. An example is given in Table \ref{tab:ex-query}.

\begin{table*}[t]
\resizebox{\textwidth}{!}{%
\begin{tabular}{ll|ll|l}
\toprule
\textbf{Query} & \textbf{Translated Query} & \textbf{Title} & \textbf{Translated Title} & \textbf{Recall}\\
\midrule
\underline{candado} bicicleta 
& bicycle \underline{lock} 
& \makecell[lt]{New Bicycle Vibration Code Moped \underline{Lock}\\ Bike Cycling Security Alarm Sound \underline{Lock}}
& \makecell[lt]{Nuevo c\'{o}digo de vibraci\'{o}n Bicicleta Ciclomotor alarma de seguridad\\ de bloqueo Bicicleta Ciclismo \underline{Cerradura} De Sonido}  
& 0.5 \\
\bottomrule
\end{tabular}%
}
\caption{Example for query and product title translation. `candado' is translated to `lock' in the query, but then translated back to `cerradura' in the title. The recall metric would prefer a title translation with `candado', as it was specified by the user.}
\label{tab:ex-query}
\end{table*}

The advantage of embedding feedback collection into a search task is that we can assume that users who formulate a search query have a genuine intent of finding products that fit their need, and are also likely to be satisfied with product title translations that match their query, i.e., contain terms from the query in their own language. We exploit this assumption in order to measure the quality of a product title translation by requiring a user to click on the translation when it is displayed as a result of the search, and then quantifying the quality of the clicked translation by the extent it matches the query that led the user to the product. For this purpose, we define a word-based matching function $\text{match}(w, \textbf{q})$ that evaluates whether a query $\mathbf{q}$ contains the word $w$:
\begin{align}  \label{eq:match}
\text{match}(w, \mathbf{q}) = \begin{cases} 
1, & \text{if} \, w \in \mathbf{q} \\
0, & \text{otherwise}. 
\end{cases}
\end{align}
Based on this word-level matching, we compute a sequence-level reward for a sentence $\mathbf{y}$ of length $T$ as follows:
\begin{align}\label{eq:recall}
\text{recall}(\mathbf{y}, \mathbf{q}) = \frac{1}{T} \sum_{t=1}^T \text{match}(y_t, \mathbf{q}).
\end{align}

\section{Learning from User Feedback}\label{sec:learning}

\paragraph{Reward Functions.}
In reinforcement and bandit learning, rewards received from the environment are used as supervision signals for learning. In our experiments, we investigate several options to obtain a reward function $\Delta: \mathcal{Y} \rightarrow [0, 1]$ from logged human bandit feedback:
\begin{enumerate}
\item \textbf{Direct User Reward}: Explicit feedback, e.g., in the form of star ratings, can directly be used as reward by treating the reward function as a black box. Since human feedback is usually only available for one translation per input, learning from direct user rewards requires the use of bandit learning algorithms. In our setup, human bandit feedback has been collected for translations of a historic MT system different from the target system to be optimized. This restricts the learning setup to offline learning from logged bandit feedback.
\item \textbf{Reward Scoring Function}: A possibility to use human bandit feedback to obtain rewards for more than a single translation per input is to score translations either against a logged reference or a logged query. The first option requires a bandit-to-supervised conversion of data where high-quality logged translations are used as references against which BLEU or other MT quality metrics can be measured. The second option uses logged queries to obtain a matching score as in Equation \ref{eq:recall}.
\item \textbf{Estimated Reward}: Another option to extend bandit feedback to all translations is to learn a parametric model of rewards, e.g., by optimizing a regression objective. The reward function is known, but the model parameters need to be trained based on a history of direct user rewards or by evaluations of a reward scoring function. 
\end{enumerate}

In the following, we present how rewards can be integrated in various objectives for NMT training.

\paragraph{Maximum Likelihood Estimation by Bandit-to-Supervised Conversion.} Most commonly, NMT models are trained with \emph{Maximum Likelihood Estimation} (MLE, Equation \ref{eq:mle-loss}) on a given parallel corpus of source and target sequences $D = \{(\mathbf{x}^{(s)}, \mathbf{y}^{(s)})\}^{S}_{s=1}$
\begin{align} \label{eq:mle-loss}
L^{\text{MLE}}(\theta) = \sum_{s=1}^S \log p_{\theta}(\mathbf{y}^{(s)}|\mathbf{x}^{(s)}). 
\end{align} 
The MLE objective requires reference translations and is agnostic to rewards. However, in a \emph{bandit-to-supervised conversion}, rewards can be used to filter translations to be used as pseudo-references for MLE training. We apply this scenario to explicit and implicit human feedback data in our experiments.

\paragraph{Reinforcement Learning by Minimum Risk Training.} When rewards can be obtained for several translations per input instead of only for one as in the bandit setup, by using a reward estimate or scoring function, \emph{Minimum Risk Training} (MRT, Equation \ref{eq:mrt-loss}) can be applied to optimize NMT from rewards. 
\begin{align}  \label{eq:mrt-loss}
R^{\text{MRT}}(\theta) = \sum_{s=1}^S  \sum_{\mathbf{\tilde{y}} \in \mathcal{S}(\mathbf{x}^{(s)})} q^{\alpha}_{\theta}(\mathbf{\tilde{y}}|\mathbf{x}^{(s)})\, \Delta (\mathbf{\tilde{y}}),
\end{align}
where sample probabilities are renormalized over a subset of translation samples $\mathcal{S}(\mathbf{x}) \subset \mathcal{Y}(\mathbf{x})$:
$q^{\alpha}_{\theta}(\mathbf{\tilde{y}}|\mathbf{x}) = \frac{p_{\theta}(\mathbf{\tilde{y}}|\mathbf{x})^{\alpha}}{\sum_{\mathbf{y}' \in \mathcal{S}(\mathbf{x})} p_{\theta}(\mathbf{y}'|\mathbf{x})^{\alpha}}$. The hyper-parameter $\alpha$ controls the sharpness of $q$ (see \citet{shen2016minimum}). 

With sequence-level rewards, all words of a translation of length $T$ are reinforced to the same extent and are treated as if they contributed equally to the translation quality. A word-based reward function, such as the match with a given query (Equation \ref{eq:match}), allows the words to have individual weights. 
The following modification of the sequence-level MRT objective (Equation \ref{eq:mrt-loss}) accounts for word-based rewards $\Delta(y_t)$:
\begin{align} \label{eq:wm}
R^{\text{W-MRT}}(\theta) =& \sum_{s=1}^S   \sum_{\mathbf{\tilde{y}} \in \mathcal{S}(\mathbf{x}^{(s)})}\prod_{t=1}^T \nonumber \\ 
& \biggl[  
q^{\alpha}_{\theta}(\tilde{y}_t|\mathbf{x}^{(s)}, \mathbf{\tilde{y}}_{<t})
\, \Delta(y_t) \biggr],
\end{align} 
where $\Delta(y_t)$ in our experiments is a matching score  \eqref{eq:match}. In the following we use the bracketed prefix (W-) to subsume both sentence-level and word-level training objectives. 

When output spaces are large and reward functions sparse, (W-)MRT objectives typically benefit from a warm start, i.e., pre-training with MLE. Following \citet{wu2016google}, we furthermore adopt a linear combination of MLE and (W-)MRT to stabilize learning:
\begin{align*}
R^{\text{(W-)MIX}}(\theta) = \lambda \cdot R^{\text{MLE}}(\theta) + R^{\text{(W-)MRT}}(\theta).
\end{align*} 

\paragraph{Counterfactual Learning by Deterministic Propensity Matching.} Counterfactual learning attempts to improve a target MT system from a log of source sentences, translations produced by a historic MT system, and obtained feedback $L = \{(\mathbf{x}^{(h)}, \mathbf{y}^{(h)}, \Delta(\mathbf{y}^{(h)}))\} ^H_{h=1}$. For the special case of deterministically logged rewards
\citet{lawrence-sokolov-riezler:2017:EMNLP2017} introduced the \emph{Deterministic Propensity Matching} (DPM) objective with self-normalization as a multiplicative control variate \cite{SwaminathanJoachimsNIPS:15}:\footnote{\citet{lawrence-sokolov-riezler:2017:EMNLP2017} propose reweighting over the whole log, but this is infeasible for NMT. For simplicty we refer to their DPM-R objective as DPM, and DC-R as DC.} 
\begin{align} \label{eq:dpm-loss}
R^{\text{DPM}}(\theta) = \frac{1}{H}\sum_{h=1}^H \Delta(\mathbf{y}^{(h)}) \, \bar{p}_{\theta}(\mathbf{y}^{(h)}|\mathbf{x}^{(h)}),
\end{align} 
where translation probabilities are reweighted over the current mini-batch $B \subset H, B \ll H$: $\bar{p}_{\theta}(\mathbf{y}^{(h)}|\mathbf{x}^{(h)}) =
\frac{p_{\theta}(\mathbf{y}^{(h)}|\mathbf{x}^{(h)})}{\sum_{b=1}^B p_{\theta}(\mathbf{y}^{(b)}|\mathbf{x}^{(b)})}$. We additionally normalize the log probability of a translation $\mathbf{y}$ by its length $|\mathbf{y}|$:
$p^{norm}_{\theta}(\mathbf{y}|\mathbf{x})
= \exp{(\frac{\log p_{\theta}(\mathbf{y}|\mathbf{x})}{|\mathbf{y}|})}.$ 

\paragraph{Counterfactual Learning by Doubly Controlled Estimation.} \citet{lawrence-sokolov-riezler:2017:EMNLP2017} furthermore propose the \emph{Doubly Controlled} objective (DC, Equation \ref{eq:dc-loss}) implementing the idea of doubly robust estimation \cite{dudik2011doubly, JiangLi:16} for deterministic logs.
In addition to learning from the historic reward for the logging system, the reward for other translations is estimated by a parametrized regression model that is trained on the log $\hat{\Delta}_{\phi}: \mathcal{Y} \rightarrow [0,1]$. This objective contains both a multiplicative (probability reweighting) and an additive (reward estimate) control variate, hence the name.\footnote{We find empirically that estimating $\hat{c}$ over the current batch as in objective $\hat{c}$DC in \cite{lawrence-sokolov-riezler:2017:EMNLP2017} does not improve over the simple setting with $c=1$.}
\begin{align}  \label{eq:dc-loss}
&R^{\text{DC}}(\theta) = \frac{1}{H}\sum_{h=1}^H \biggl[ \left( \Delta(\mathbf{y}^{(h)}) - \hat{\Delta}_{\phi}(\mathbf{y}^{(h)}) \right)  \nonumber \\ 
 & \times \bar{p}_{\theta}(\mathbf{y}^{(h)}|\mathbf{x}^{(h)}) +\sum_{\mathbf{y} \in \mathcal{S}(\mathbf{x}^{(h)})} \hat{\Delta}_{\phi}(\mathbf{y}) \, p_{\theta} (\mathbf{y}|\mathbf{x}^{(h)}) \biggr] 
\end{align}
As for MRT, the expectation over the full output space is approximated with a subset of $k$ sample translations $\mathcal{S}(\mathbf{x}) \subset \mathcal{Y}(\mathbf{x})$.

\paragraph{Relative Rewards.}
With the objectives as defined above, gradient steps are dependent on the magnitude of the reward for the current training instance. In reinforcement learning, an average reward baseline is commonly subtracted from the current reward with the primary goal to reduce variance \cite{williams1992simple}. As a side effect, the current reward is relativized, such that the gradient step is not only determined by the magnitude of the current rewards, but is put into relation with previous rewards. We found this effect to be particularly beneficial in experiments with suboptimal reward estimators or noisy rewards and therefore apply it to all instantiations of the DPM and DC objectives. For DPM, the running average of historic rewards $\bar{\Delta}_h= \frac{1}{h} \sum ^h_{i=1}\Delta(\mathbf{y}^{(i)})$ is subtracted from the current reward. For DC we apply this to both types of rewards in Equation \ref{eq:dc-loss}: 1) the logged reward $\Delta(\mathbf{y}^{(h)})$, from which we subtract its running average $\bar{\Delta}_h$ instead of the estimated reward $\hat{\Delta}_{\phi}(\mathbf{y}^{(h)})$, and 2) the estimated reward $\hat{\Delta}_{\phi}(\mathbf{y})$, from which we hence subtract the average estimated reward $\bar{\hat{\Delta}}_h= \frac{1}{h} \sum ^h_{i=1} \frac{1}{k} \sum_{\mathbf{y'} \in \mathcal{S}(\mathbf{x}^{(i)})} \hat{\Delta}_{\phi}(\mathbf{y'})$.

\section{Experiments}\label{sec:exps}

\subsection{NMT Model}
In our experiments, learning from feedback starts from a pre-trained English to Spanish NMT model that has not seen in-domain data (i.e., no product title translations). The NMT baseline model (BL) is a standard subword-based encoder-decoder architecture with attention \cite{BahdanauETAL:15}, implemented with TensorFlow \cite{tensorflow2015-whitepaper}. The model is trained with MLE on 2.7M parallel sentences of out-of-domain data until the early stopping point which is determined on a small in-domain dev set of 1,619 product title translations. A beam of size 12 and length normalization \cite{wu2016google} are used for beam search decoding. For significance tests we used approximate randomization \cite{clark-EtAl:2011:ACL-HLT2011}, for BLEU score evaluation (lowercased) the multi-bleu script of the Moses decoder \cite{koehn2007moses},
for TER computation the tercom tool \cite{snover2006study}. For MRT, DC and (W-)MIX models we set $k=5$, for (W-)MIX models $\lambda=0.5$ and $\alpha=0.05$. For all NMT models involving random sampling, we report average results and standard deviation (in subscript) over two runs. Further details about training data and hyperparameters settings are described in Appendix \ref{sec:traindetails}. 

\subsection{Reward Estimator}\label{sec:re}
\begin{figure}
\includegraphics[width=\columnwidth]{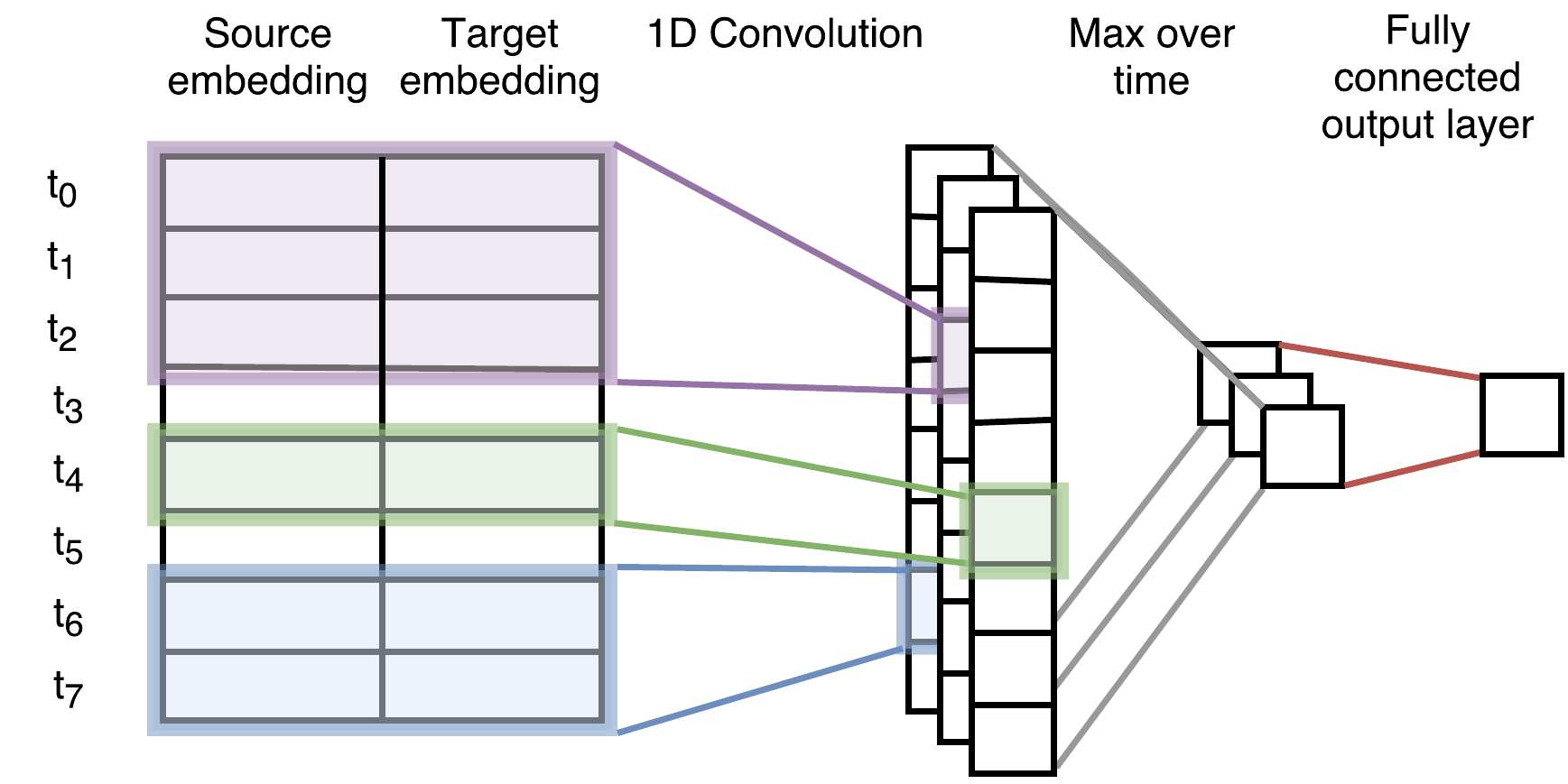}
\caption{Model architecture for the reward estimator. This example has one filter for each filter size (3: purple, 1: green, 2: blue). Source and target sequences are padded up to a maximum length, here $T_{max}=8$.}
\label{fig:cnn-re}
\end{figure}

The model architecture for the reward estimator used in the DC objective is a bilingual extension of the convolutional neural network (CNN) for sentence classification proposed by \citet{kim:2014:EMNLP2014}. Both source and target sequences are padded up to a pre-defined maximum sequence length $T_{max}$, their embeddings are concatenated and further processed by a 1D-Convolution over the time dimension with several filters of sizes from 2 to 15, which is then followed by a max-over-time pooling and fed to a fully-connected output layer (Figure \ref{fig:cnn-re}).  
The model is trained to minimize the mean squared error (MSE) on the training portion of the logged feedback data (60k for simulated sentence-BLEU feedback, 62,470 for star rating feedback). 
The word embeddings of the reward estimator are initialized by the word embeddings of the trained baseline NMT system and fine-tuned further together with the other CNN weights. The best parameters are identified by early-stopping on the validation portion of the feedback data (2,162 for the simulation, 6,942 for the star ratings). Please find a detailed description of the model's hyperparameters in Appendix \ref{sec:redetails}.

Results for a stand-alone evaluation of the reward estimator on the validation portions of the feedback data are given in Table \ref{tab:re}. The estimator models sBLEU much more accurately than the user star ratings. This is due to large variance and skew of the user ratings. An MSE-trained estimator typically predicts values around the mean, which is not a suitable strategy for such a skewed distribution of labels, but is successful for the prediction of normal-distributed sBLEU. 

\begin{table}[t]
\centering
\resizebox{\columnwidth}{!}{%
\begin{tabular}{l|c|c|c|c|c}
\toprule
\textbf{Data \& Model} & \textbf{MSE} & \thead{\textbf{Macro-avg.}\\ \textbf{Distance}} & \thead{\textbf{Micro-Avg.} \\ \textbf{Distance}} & \thead{\textbf{Pearson's}\\ \textbf{r}} & \thead{\textbf{Spearman's}\\ $\mathbf{\rho}$}\\
\midrule
Star ratings & 0.1620 & 0.0065 & 0.3203 & 0.1240 & 0.1026\\
sBLEU & 0.0096 & 0.0055 & 0.0710 & 0.8816 & 0.8675 \\
\bottomrule
\end{tabular}%
}
\caption{Results for the \textbf{reward estimators} trained and evaluated on human star ratings and simulated sBLEU.}
\label{tab:re}
\end{table}

\subsection{Explicit Star Rating Feedback}

\paragraph{Counterfactual Bandit Learning.}
As shown in Table \ref{tab:user}, counterfactual learning with DPM and DC on the logged star ratings as direct reward does not yield improvements over the baseline model in terms of corpus BLEU or TER. A randomization of feedback signals for translations gives the same results (DPM-random), showing that counterfactual learning from logged star ratings is equivalent to learning from noise.
Evaluating the models in terms of estimated user reward, however, we find an improvement of +1.49 for DC, +0.04 for DPM over the baseline (53.93) (not shown in Table \ref{tab:user})---but these improvements do not transfer to BLEU because the reward model largely over-estimates the translation quality of translations with major faults.
Hence it is not desirable to optimize towards this signal directly. 

\paragraph{Bandit-to-Supervised Conversion.}
In the following setup, we utilize the user ratings to filter the log by using only five star rated translations, and perform supervised learning of MLE and  MIX using sBLEU against pseudo-references as reward function. Table \ref{tab:user} shows that this filtering strategy leads to large improvements over the baseline, for MLE and even more for MIX, even though the data set size is reduced by 42\%. However, around the same improvements can be achieved with a random selection of logged translations of the same size (MIX small, containing 55\% five-star ratings). Using all logged translations for training MIX achieves the best results. This suggests that the model does not profit from the feedback, but mostly from being exposed to in-domain translations of the logging system. This effect is similar to training on pseudo-references created by back-translation \cite{sennrich-haddow-birch:2016:P16-11, sennrich-haddow-birch:2016:WMT}.

\begin{table}[t]
\centering
\resizebox{0.7\columnwidth}{!}{%
\begin{tabular}{l|c|c}
\toprule
\textbf{Model} & \textbf{Test BLEU} & \textbf{Test TER} \\
\midrule
BL & 28.38 & 57.58 \\
\midrule
 DPM &  28.19& 57.80\\
 DPM-random & 28.19 & 57.64\\
 DC & 28.41$_{\pm 0.85}$  & 64.25$_{\pm 1.66}$ \\
\midrule
 MLE (all) &  31.98 & 51.08\\
 MIX (all) & \textbf{34.47}$_{\pm 0.06}$ & \textbf{47.97}$_{\pm 0.18}$ \\ 
 MIX (small) & 34.16$_{\pm 0.09}$ & 48.12$_{\pm 0.33}$\\ 
 MIX ($\text{stars}=5$) & 34.35$_{\pm 0.11}$ & 47.99$_{\pm 0.13}$\\ 
\bottomrule
\end{tabular}%
}
\caption{Results for models trained on \textbf{explicit user ratings} evaluated on the product titles test set. 
`small' indicates a random subset of logged translations of the same size as the filtered log that only contains translations with an average rating of five stars (`$\text{stars}=5$'). The differences in BLEU  are not significant at $p\leq0.05$ between MIX models, but over other models.}
\label{tab:user}
\end{table}

\subsection{Task-Based Implicit Feedback}

\paragraph{Bandit-to-Supervised Conversion.}
We apply the same filtering technique to the logged implicit feedback by treating translations with $\text{recall}=1$ as references for training MIX with sBLEU (reduction of the data set by 62\%). The results in Table \ref{tab:query} show that large improvements over the baseline can be obtained even without filtering, BLEU and TER scores being comparable to the ones observed for training on explicit user ratings.

\begin{table}[t]
\centering
\resizebox{0.7\columnwidth}{!}{%
\begin{tabular}{l|c|c}
\toprule
\textbf{Model} & \textbf{Test BLEU} & \textbf{Test TER}\\
\midrule
BL & 28.38 & 57.58 \\
\midrule
 MLE (all) & 31.89 & 51.35\\
 MIX (all) & 34.39$_{\pm0.08}$ & 47.94$_{\pm0.24}$\\ 
 MIX (small) & 34.13$_{\pm0.26}$ & 48.27$_{\pm0.60}$\\  
 MIX ($\text{recall}=1$) & 34.17$_{\pm0.02}$ & 47.72$_{\pm0.26}$\\ 
\midrule
W-MIX & \textbf{34.52}$_{\pm 0.02}$ & \textbf{46.91}$_{\pm 0.03}$\\
\bottomrule
\end{tabular}%
}
\caption{Results for models trained on \textbf{implicit task-based} feedback data evaluated on the product titles test set. 
`small' indicates a random subset of logged translations of the same size as the filtered log that only contains translations that contain all the query words (`$\text{recall}=1$'). The BLEU score of MIX (small) significantly differs from MIX (all) at $p\leq0.05$, the score of MIX ($\text{recall}=1$) does not. Other differences are significant.}
\label{tab:query}
\end{table}

\paragraph{Task-based Feedback.}
The key difference between the implicit feedback collected in the query-title data and the explicit user ratings, is that it can be used to define reward functions like recall or match (Equations \ref{eq:recall}, \ref{eq:match}). For the experiments we train W-MIX, the word-based MRT objective (Equation \ref{eq:wm}) linearly combined with MLE, on the logged translations accompanying the queries (160k sentences).
This combination is essential here, since the model would otherwise learn to produce translations that contain nothing but the query words.
To account for user-generated language in the queries and subwords in the MT model, we soften the conditions for a match, counting tokens as a match that are part of a word $w$ that is either contained in the query, or has edit distance to a word in the query with $dist(w,\mathbf{q}_i) < \max(3, 0.3\times|w|)$.

\begin{table}[t]
\centering
\resizebox{\columnwidth}{!}{%
\begin{tabular}{l|c|c|c|c}
\toprule
 \textbf{Logged} & \textbf{BL} & \textbf{MIX (Tab. \ref{tab:user})} & \textbf{MIX (Tab. \ref{tab:query})} & \textbf{W-MIX}\\
\midrule
65.33 & 45.96 &  62.92$_{\pm0.56}$ & 63.21$_{\pm0.24}$ & \textbf{68.12}$_{\pm0.27}$ \\
\bottomrule
\end{tabular}%
}
\caption{\textbf{Query recall} results on the query test set, comparing the logged translations, the baseline and the best MIX models trained on logged translations (MIX (all) from Tables \ref{tab:user} and \ref{tab:query}) with the W-MIX model trained via word-based query matching (W-MIX from Table \ref{tab:query}). 
}
\label{tab:queryrecall}
\end{table}

Table \ref{tab:queryrecall} repeats the best MIX results from Table 
\ref{tab:user} and \ref{tab:query}, and evaluates the models with respect to query recall. We also report the query recall for the logged translations and the out-of-domain baseline. These results are compared to W-MIX training on implicit feedback data described in Section \ref{sec:implicit}. The development portion of the query-title dataset contains 4,065 sentences, the test set 2,000 sentences, which is used for query recall evaluation. 
The W-MIX model shows the largest improvement in query recall (12\% points) and BLEU (6 points) over the baseline out of all tested learning approaches. It comes very close to the BLEU/TER results of the model trained on in-domain references, but surpasses its recall by far. This is remarkable since the model does not use any human generated references, only logged data of task-based human feedback. Appendix \ref{sec:examples} contains a set of examples illustrating what the W-MIX learned.

\section{Conclusion}
We presented methods to improve NMT from human reinforcement signals. The signals were logged from user activities of an e-commerce platform and consist of explicit ratings on a five-point Likert scale and implicit task-based feedback collected in a cross-lingual search task. We found that there are no improvements when learning from user star ratings, unless the noisy ratings themselves are stripped off in a bandit-to-supervised conversion. 
Implicit task-based feedback can be used successfully as a reward signal for NMT optimization, leading to improvements both in terms of enforcing individual word translations and in terms of automatic evaluation measures. 
In the future, we plan transfer these findings to production settings by performing regular NMT model updates with batches of collected user behavior data, especially focusing on improving translation of ambiguous and rare terms based on rewards from implicit partial feedback. 

\section*{Acknowledgements}
The last author was supported in part by DFG Research Grant RI 2221/4-1. We would like to thank Pavel Petrushkov for helping with the NMT setup, and the anonymous reviewers for their insightful comments .

\bibliography{naaclhlt2018}
\bibliographystyle{acl_natbib}

\clearpage

\appendix
\section{Appendix Overview}
Section \ref{sec:instructions} provides the instructions that were given to the annotators when judging MT quality. In Section \ref{sec:rewards} we provide histograms for simulated and explicit rewards.
Section \ref{sec:traindetails} contains details on the data and NMT model hyperparameters. In Section \ref{sec:simulations} we give results for simulation experiments on the e-commerce product title domain and a publicly available data set. Finally,  we compare translation examples of different models in Section \ref{sec:examples}.

\section{Annotation Instructions}\label{sec:instructions}
\subsection{Star Ratings}\label{sec:rating-instruct}
Please rate the translation quality of the segments on the scale from 1 to 5. 
Focus on whether or not the information contained in the source sentence is correctly and completely translated (ratings 1 - 4). Then, if you are ready to give a 4 based on the criteria below, check whether or not you can assign a 5 instead of the 4, focusing on remaining grammatical, morphological and stylistic errors. Remember that even a very fluent translation that looks like a human-produced sentence can receive a bad rating if it does not correctly convey all the information that was present in the source.

Assign the following ratings from 1 to 5:
\begin{enumerate}
\item Important information is missing and/or distorted in the translation, and the error is so severe that it may lead to erroneous perception of the described product. Or the translation contains profanities/insulting words. 
\item Information from the source is partially present in the translation, but important information is not translated or translated incorrectly. 
\item The most important information from the source is translated correctly, but some other less important information is missing or translated incorrectly. 
\item All of the information from the source is contained in the translation. This should be the only criterion to decide between 1-3 and 4. It is okay for a 4-rated translation to contain grammatical errors, disfluencies, or word choice that is not very appropriate to the style of the input text. There might be errors in casing of named entities when it is clear from the context that these are named entities.
\item All of the information from the source is contained in the translation and is translated correctly. In contrast to a 4-rated translation, the translation is fluent, easy to read, and contains either no or very minor grammatical/morphological/stylistic errors. The brand names and other named entities have the correct upper/lower case.
\end{enumerate}

\subsection{Binary Judgment}
The customers of the eBay e-commerce platform, when presented with a title translation on the product page, can hover with the mouse over the translation of the title and see the original (source) title in a pop-up window. There, they have the possibility to rate the translation with 1 to 5 stars.

The goal of this evaluation is to check the ratings - you have to mark ``Agree'' when you agree with the rating and ``Disagree'' otherwise. The rating (number from 1 to 5) is shown in the Reference line.

Note that eBay customers did not have any instructions on what the rating of 5 stars, 3 stars, or 4 stars means. Thus, the evaluation is subjective on their side.
Please apply your common sense when agreeing or disagreeing with human judgment. The focus should be on adequacy (correct information transfer) as opposed to fluency.

\section{Rewards}\label{sec:rewards}
\subsection{Reward Distributions}
\begin{figure}[h]
\includegraphics[width=\columnwidth]{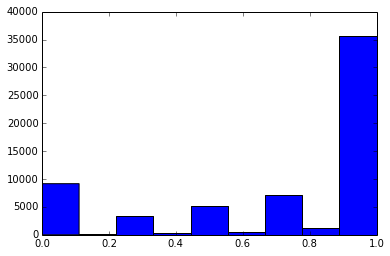}
\caption{Distribution of user star ratings. The original ratings on a five-star scale are averaged per title and rescaled.}
\label{fig:ratings}
\end{figure}

\begin{figure}[h]
\includegraphics[width=\columnwidth]{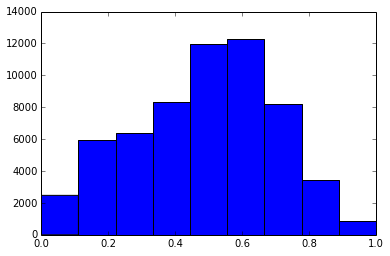}
\caption{Distribution of sentence BLEUs of the product title training set when translated with the out-of-domain baseline for simulation experiments.}
\label{fig:sbleu}
\end{figure}

Figure \ref{fig:ratings} shows the distribution of logged user star ratings, Figure \ref{fig:sbleu} the distribution of sentence BLEU (sBLEU) scores for the simulation experiments with logged feedback. The logged translations for the user star ratings were generated by the production system, the logged translations for the simulation were generated by the BL NMT system.

\section{Training Details}\label{sec:traindetails}
\subsection{Data}

We conduct experiments on an English-to-Spanish e-commerce item titles translation task.
The in-domain data for training with simulated feedback is composed of in-house eBay data (item titles, descriptions, etc.). The out-of-domain data for training the baselines contains only publicly available parallel corpora, that is Europarl, TAUS, and OpenSubtitles released by the OPUS project~\cite{tiedemann2009news}. 
The out-of-domain data has been sub-sampled according to the similarity to the domain of the product title data, and 25\% of the most similar sentence pairs have been selected.
The corpus statistics for parallel data are shown in Table \ref{tab:data}. Before calculating the corpus statistics, 
we apply pre-processing including tokenization and replacement of
numbers and product specifications with a placeholder token (e.g., `6S', and `1080p').
Table \ref{tab:data-2} gives 
an overview of the type and the size of the translations with feedback. 

\begin{table}[h]
\begin{center}
\resizebox{0.9\columnwidth}{!}{%
\begin{tabular}{lr|c|c}
\toprule
       &              &  \textbf{En}    &   \textbf{Es}   \\ 
       \midrule
Train: & Sentences    & \multicolumn{2}{c}{2,741,087} \\               
       & Tokens       & 17,139,416 & 18,270,753\\
       & Vocabulary   & 327,504 & 393,757\\ 
       & Singletons   & 162,834 & 190,686\\ 
       
\midrule
Dev.:  & Sentences    & \multicolumn{2}{c}{1,619}\\ 
       & Tokens       & 29,063 & 31,813\\ 
       & Vocabulary   & 11,314 & 11,532\\
	   & OOVs         & 2,645 & 2,493\\
       
\midrule
Test   & Sentences    & \multicolumn{2}{c}{1000} \\ 
       & Tokens       & 9,851 & 11,221 \\
       & Vocabulary   & 6,735 & 6,668 \\ 
       & OOVs         & 1,966 & 1,902\\     
\bottomrule
\end{tabular}%
}
\caption{Corpus statistics for the out-of domain training data and in-domain dev and test data.}
\label{tab:data}
\end{center}
\end{table}

\begin{table}[h]
\centering
\resizebox{0.6\columnwidth}{!}{%
\begin{tabular}{l|c}
\toprule
\textbf{Description} & \textbf{Size} \\
\midrule
User star ratings & 69,412 \\
$\dots$ with 5 stars & 40,064 \\
\midrule
Expert star ratings & 1,000\\
Expert judgments & 2,000\\
\midrule
Query-title pairs & 164,065\\
$\dots$ with $recall=1$ & 61,965 \\
\midrule
Title translations & 62,162 \\
\bottomrule
\end{tabular}%
}
\caption{Data set sizes for collected feedback in number of sentences. The in-domain title translations are only used for simulation experiments.}
\label{tab:data-2}
\end{table}

\subsection{NMT Model Architecture}
The NMT has a bi-directional RNN encoder with one layer of 1000 GRUs, a decoder with 1000 GRUs, and source and target word embeddings of size 620. The vocabulary is generated from the out-of-domain training corpus with 40k byte-pair merges \cite{sennrich-haddow-birch:2016:P16-12} and contains 40813 source tokens and 41050 target tokens. The full softmax is approximated by 1024 samples as proposed in \cite{jean-EtAl:2015:ACL-IJCNLP}. Dropout \cite{gal2016dropout} is applied with probability $p=0.1$ to the embedding matrices, with $p=0.2$ to the input and recurrent connections of the RNNs. 

\subsection{NMT Training Hyperparameters}
The out-of-domain model is trained with minibatches of size 100 and L2 regularization with weight $\num{1e-7}$, optimized with Adam \cite{kingma2014adam} with initial $\alpha=0.0002$, then decaying $\alpha$ by 0.9 each epoch.

The remaining models are trained with constant learning rates and mini-batch size 30, regularization and dropout stay the same. The settings for the other hyperparameters are listed in Table \ref{tab:hyperparam}. The estimator loss weight is only relevant for DC, where the pre-trained estimator gets further fine-tuned during DC training.

\begin{table*}[h]
\resizebox{\textwidth}{!}{%
\begin{tabular}{l|cccccc}
\toprule
\textbf{Model} & \textbf{Adam's} $\alpha$  & \textbf{Length-Normalization} & \textbf{MRT} $\alpha$ & \textbf{Sample Size} $k$ & \textbf{MIX} $\lambda$ & \textbf{Estimator Loss Weight}\\
\midrule
\multicolumn{7}{l}{Simulated Feedback} \\
\midrule
MLE & 0.002 &  - & - & - & - & -\\
MIX & 0.002 & - & 0.005 & 5 & 0.05 &- \\
EL & \num{2e-6} &   - & - & - & - & -\\
DPM & \num{2e-6} & x & - & - & - & -\\
DPM-random & \num{2e-6}  & x & - & - & - & - \\
DC & 0.002  & - & - & 5 & - & 1000\\
\midrule
\multicolumn{7}{l}{Explicit Star Rating Feedback} \\
\midrule
DPM & \num{2e-6}  & x & - & - & - & -\\
DPM-random & \num{2e-6}  & x & - & - & - & -\\
DC & \num{2e-6} & x & - & 5 & - & 1000 \\
MLE (all) & 0.002 & - & - & - & - & -\\
MIX (all) & 0.002  & - &  0.005 & 5 & 0.05 &- \\
MIX (small) & 0.002  & - &  0.005 & 5 & 0.05 &- \\
MIX (stars=5) & 0.002  & - &  0.005 & 5 & 0.05  &-\\
\midrule
\multicolumn{7}{l}{Implicit Task-Based Feedback } \\
\midrule
MLE (all) &  0.002  & -  & - & - & - & - \\
MIX (all) & 0.002 & - &  0.005 & 5 & 0.05 &- \\
MIX (small) & 0.002  & - &  0.005 & 5 & 0.05 &-  \\
MIX (recall=1) & 0.002  & - &  0.005 & 5 & 0.05 &- \\
W-MIX & 0.002  & - &  0.005 & 5 & 0.05 &-  \\
\bottomrule
\end{tabular}%
}
\caption{Hyperparameter settings for training of the models.}
\label{tab:hyperparam}
\end{table*}

\subsection{Reward Estimation} \label{sec:redetails}
We find that for reward estimation a shallow CNN architecture with wide filters performs superior to a deeper CNN architecture \cite{le:2017:arxiv} and also to a recurrent architecture. Hence, we use one convolutional layer with ReLU activation of $n_{f}$ filters each for filter sizes from 2 to 15, capturing both local and more global features. For reward estimation on star ratings, $n_{f}=100$ and on simulated sBLEU $n_{f}=20$ worked best. Dropout with $p=0.5$ is applied before the output layer for the simulation setting. We set $T_{max}=60$.
The loss of each item in the batch is weighted by inverse frequency of its feedback in the current batch (counted in 10 buckets) to counterbalance skewed feedback distributions. The model is optimized with Adam \cite{kingma2014adam} (constant $\alpha=0.001$ for star ratings, $\alpha=0.002$ for the simulation) on minibatches of size 30.
Note that the differences in hyper-parameters between both settings are the result of tuning and do not cause the difference in quality of the resulting estimators. We do not evaluate on a separate test set, since their final quality can be measured in how much well they serve as policy evaluators in counterfactual learning.

\section{Simulated Bandit Feedback} \label{sec:simulations}
\paragraph{Expected Loss.} When rewards can be retrieved for sampled translations during learning, the Online Bandit Structured Prediction framework proposed by \citet{sokolov-EtAl:2016:P16-1, sokolovNIPS2016} can be applied for NMT, as demonstrated in \citet{kreutzer-sokolov-riezler:2017:Long, sokolov-EtAl:2017:WMT}. The \emph{Expected Loss} objective (EL, Equation \ref{eq:el-loss}) maximizes\footnote{We use the terms reward or loss interchangeably depending on minimization or maximization contexts.} the expectation of a reward over all source and target sequences, and does in principle not require references:
\begin{align} \label{eq:el-loss}
R^{\text{EL}}(\theta) =& \mathbb{E}_{p(\mathbf{x}) p_{\theta}(\mathbf{\tilde{y}}|\mathbf{x})} \left[\Delta (\mathbf{\tilde{y}}) \right].
\end{align} 
While we could not apply it to the logged user feedback since it was obtained offline, we can compare to its performance in a simulation setting with simulated rewards instead of human feedback. It is expected to outperform methods learning with logged feedback due to the exploration during learning.
In the following simulation experiments, $\Delta (\mathbf{\tilde{y}})$ is computed by comparing a sampled translation $\mathbf{\tilde{y}} \sim p_{\theta}(\mathbf{y}|\mathbf{x})$ to a given reference translation $\mathbf{y}$ with smoothed sentence-level BLEU (sBLEU).

\subsection{E-commerce Product Titles}
We test several of the proposed learning techniques with an in-domain parallel corpus (62,162 sentences) of product titles where bandit feedback is simulated by evaluating a sampled translation against a reference using sBLEU. Similar to previous studies on SMT \cite{lawrence-sokolov-riezler:2017:EMNLP2017}, this reward is deterministic and does not contain user-dependent noise. 

\paragraph{Supervised Fine-Tuning.}
When fine-tuning the baseline model on in-domain references \cite{luong2015stanford}, the model improves 3.34 BLEU (MLE in Table \ref{tab:simulation}) on an in-domain test set (1,000 sentences). By tuning it on the same in-domain data for sBLEU with MIX, it gains another 3 BLEU points. 

\paragraph{Bandit Learning.} When feedback is given to only one translation per input (=online bandit feedback), the model (EL) achieves comparable performance to MLE training with references. When the feedback is logged offline for one round of deterministic outputs of the baseline model (=offline bandit feedback), we can still find improvements of 1.81 BLEU (DPM). With a reward estimator trained on this log, DC achieves even higher improvements of 3 BLEU. To test the contribution of the feedback in contrast to a simple in-domain training effect, we randomly perturbed the pairing of feedback signal and translation and retrain (DPM-random). This clearly degrades results, confirming feedback to be a useful signal rather than noise.
\begin{table}[t]
\centering
\resizebox{\columnwidth}{!}{%
\begin{tabular}{ll|c|c}
\toprule
\textbf{Learning} & \textbf{Model}  & \textbf{Test BLEU} & \textbf{Test TER}\\
\midrule
Pre-trained &BL &  28.38 & 57.58 \\
\midrule
\multirow{2}{*}{Fully Supervised} & MLE  & 31.72 & 53.02\\
& MIX  & 34.79$_{\pm 0.02}$ & 48.56$_{\pm 0.02}$ \\ 
\midrule
Online Bandit& EL  & 31.78$_{\pm 0.06}$ & 51.11$_{\pm 0.36}$ \\
\midrule
\multirow{3}{*}{Counterfactual} & DPM  & 30.19 & 56.28\\
& DPM-random   & 28.20 & 57.89\\
& DC & 31.11$_{\pm 0.34}$ & 55.05$_{\pm 0.02}$\\
\bottomrule
\end{tabular}%
}
\caption{Results for \textbf{simulation experiments} evaluated on the product titles test set. 
}
\label{tab:simulation}
\end{table}

\begin{table}[t]
\centering
\resizebox{\columnwidth}{!}{%
\begin{tabular}{l|c|c|c}
\toprule
\textbf{Model} & \textbf{SMT} & \textbf{NMT} (beam search) & \textbf{NMT} (greedy)\\
\midrule
EP BL & 25.27 & 27.55 & 26.32\\
NC BL & -- & 22.35 & 19.63\\
\midrule
 MLE & 28.08 & 32.48 & 31.04 \\
\midrule 
 EL & -- & 28.02 & 27.93\\
 \midrule
 DPM & 26.24 & 27.54 & 26.36 \\
 DC & 26.33 &28.20 & 27.39\\
\bottomrule
\end{tabular}%
}
\caption{BLEU results for simulation models evaluated on the News Commentary test set (\texttt{nc-test2007}) with beam search and greedy decoding. SMT results are from \citet{lawrence-sokolov-riezler:2017:EMNLP2017}.}
\label{tab:ep-nc}
\end{table}

\subsection{Results on Publicly Available Data}\label{sec:publicresults}
Simulation experiments were also run on publicly available data. We use the same data, preprocessing and splits as \cite{lawrence-sokolov-riezler:2017:EMNLP2017} to compare with their French-to-English news experiments on counterfactual learning with deterministically logged feedback for statistical machine translation (SMT). The baseline model is trained with MLE on 1.6M Europarl (EP) translations, bandit feedback is then simulated from 40k News Commentary (NC) translations.  
For the comparison of full supervision vs. weak feedback, we train in-domain models with MLE on in-domain NC references: training only on in-domain data (NC BL), and fine-tuning the out-of-domain baseline (EP BL) on in-domain data (MLE). The results are given in Table \ref{tab:ep-nc}. The NMT baselines outperform the SMT equivalents. With fully supervised fine-tuning the NMT models improve over the out-of-domain baseline (EP BL) by 5 BLEU points, outperforming also the in-domain baseline (NC BL). Moving to weak feedback, we still find improvements over the baseline by 0.5 BLEU with beam search and 1.6 BLEU with greedy decoding for online feedback (EL), and 0.6 BLEU with beam search and 1 BLEU with greedy decoding for counterfactual learning with DC. However, DPM performs worse than for SMT and those not manage to improve over the out-of-domain baseline. Nevertheless these results confirm that -- at least in simulation settings -- the DC objective is very suitable for counterfactual learning from bandit feedback for NMT, almost reaching the gains of learning from online bandit feedback.

\section{Examples}\label{sec:examples}

Table \ref{tab:ex1} gives an example where W-MIX training improved lexical translation choices. Table \ref{tab:ex2} lists two examples of W-MIX translations in comparison to the baseline and logged translations for given queries and product titles to illustrate the specific difficulties of the domain.

\begin{table*}[t]
\resizebox{\textwidth}{!}{%
\begin{tabular}{ll}
\toprule
Title (en) & hall linvatec pro2070 powerpro ao \underline{drill} synthes dhs \& dcs \underline{attachment} / warranty\\
\midrule
Reference-0 (es) & hall linvatec pro2070 powerpro ao \underline{taladro} synthes dhs \& dcs \underline{accesorio} / garant\'{i}a \\
Reference-1 (es) & hall linvatec pro2070 powerpro synthes , \underline{perforaci\'{o}n} , \underline{accesorio} de dhs y dcs , todo original , garant\'{i}a \\
\midrule
BL (es) & hall linvatec pro2070 powerpro ao \underline{perforadora} synthes dhs \& dcs \underline{adjuntos} / garant\'{i}a \\
\midrule
MIX on star-rated titles (es) & hall linvatec pro2070 powerpro ao \underline{perforadora} synthes dhs \& dcs \underline{adjuntos} / garant\'{i}a \\
MIX on query-titles, small (es) & hall linvatec pro2070 powerpro ao \underline{perforadora} synthes dhs \& dcs \underline{adjuntos} / garant\'{i}a \\
MIX on query-titles, all (es) & hall linvatec pro2070 powerpro ao \underline{taladro} synthes dhs \& dcs \underline{adjuntos} / garant\'{i}a
\\
W-MIX & hall linvatec pro2070 powerpro ao \underline{taladro} synthes dhs \& dcs \underline{accesorio} / garant\'{i}a \\
\bottomrule
\end{tabular}%
}
\caption{Example for product title translation from the test set where W-MIX improved the lexical choice over BL and MIX on in-domain title set and MIX on full query-title set (`perforadora' vs `taladro' as translation for `drill', `adjuntos' vs `accesorio' as translation for `attachment').}
\label{tab:ex1}
\end{table*}

\begin{table*}[t]
\resizebox{\textwidth}{!}{%
\begin{tabular}{ll}
\toprule
Title (en) & Unicorn Thread 12pcs Makeup \underline{Brushes} Set Gorgeous Colorful \underline{Foundation} Brush \\
\midrule
Query (es) & unicorn \underline{brushes} // makeup \underline{brushes} // \underline{brochas} de unicornio // \underline{brochas} unicornio \\
Query (en) & unicorn \underline{brushes} // makeup \underline{brushes} \\
\midrule
BL (es) &  \underline{galletas} de maquillaje de 12pcs \\
Log (es ) & Unicorn Rosca 12 un. Conjunto de \underline{Pinceles} para Maquillaje Hermosa Colorida \underline{Base} Cepillo \\
\midrule
W-MIX & unicornio rosca 12pcs \underline{brochas} maquillaje conjunto precioso colorido \underline{fundaci\'{o}n} cepillo
\\
\midrule
\midrule
Title (en)& 12 $\times$ Men Women Plastic Shoe Boxes 33*20*12cm Storage Organisers Clear Large Boxes \\
\midrule
Query (es) & cajas plasticas \underline{para} zapatos \\
Query (en) & plastic shoe boxes \\
\midrule
BL (es) & 12 $\times$ hombres mujeres zapatos de pl\'{a}stico cajas de almacenamiento 33*20*12cm organizadores de gran tama\~{n}o \\
Log (es) & 12 $\times$ Zapato De Hombre Mujer De Pl\'{a}stico Cajas Organizadores de almacenamiento 33*20*12cm cajas Grande Claro \\
\midrule
W-MIX & 12 $\times$ \underline{para} hombres zapatos de pl\'{a}stico cajas de pl\'{a}stico 33*20*12cm almacenamiento organizador transparente grandes cajas \\
\bottomrule
\end{tabular}%
}
\caption{Examples for product title translations of the logged query test set. In the first example, the W-MIX model improves the translation of ``brushes'', but also chooses a worse translation for ``foundation'' (``fundaci\'{o}n'' vs ``base''). In the second example, one of the tricky parts is to translate the sequence of nouns ``Men Women Plastic Shoe Boxes'' and to disambiguate the relations between them. The BL model translates ``shoes of plastic'', the Log has ``woman of plastic'' and the W-MIX model makes it ``shoes of plastic'' and ``boxes of plastic''. The W-MIX model learns to use ``para'' from the query, but omits the translation of ``women''. }
\label{tab:ex2}
\end{table*}

\end{document}